\documentclass[preprint,12pt]{elsarticle}

\usepackage[utf8]{inputenc}
\usepackage{amsmath,amssymb}
\usepackage{graphicx}
\usepackage{xcolor}
\usepackage{amsmath,amssymb,amsthm}

\usepackage{subfigure}
\usepackage{booktabs}
\usepackage[hidelinks]{hyperref}

\usepackage{array}
\usepackage{booktabs}
\newcolumntype{C}[1]{>{\centering\arraybackslash}m{#1}}

\usepackage{caption}

\journal{Engineering Applications of Artificial Intelligence}

\begin{document}

\begin{frontmatter}



\title{Reinforcement Learning for Control Systems with Time Delays: A Comprehensive Survey}

\author{Armando Alves Neto}
\affiliation{organization={Department of Electronics Eng. (DELT), UFMG},
            city={Belo Horizonte},
            country={Brazil}
            }



\begin{abstract}
In the last decade, Reinforcement Learning (RL) has achieved remarkable success in the control and decision-making of complex dynamical systems. However, most RL algorithms rely on the Markov Decision Process assumption, which is violated in practical cyber-physical systems affected by sensing delays, actuation latencies, and communication constraints. Such time delays introduce memory effects that can significantly degrade performance and compromise stability, particularly in networked and multi-agent environments.
This paper presents a comprehensive survey of RL methods designed to address time delays in control systems. We first formalize the main classes of delays and analyze their impact on the Markov property. We then systematically categorize existing approaches into five major families: state augmentation and history-based representations, recurrent policies with learned memory, predictor-based and model-aware methods, robust and domain-randomized training strategies, and safe RL frameworks with explicit constraint handling. For each family, we discuss underlying principles, practical advantages, and inherent limitations.
A comparative analysis highlights key trade-offs among these approaches and provides practical guidelines for selecting suitable methods under different delay characteristics and safety requirements. Finally, we identify open challenges and promising research directions, including stability certification, large-delay learning, multi-agent communication co-design, and standardized benchmarking. This survey aims to serve as a unified reference for researchers and practitioners developing reliable RL-based controllers in delay-affected cyber-physical systems.
\end{abstract}




\end{frontmatter}



\section{Introduction}
\label{sec:introduction}

Reinforcement Learning (RL) has emerged as a powerful paradigm for decision-making and control in complex dynamical systems, achieving remarkable performance in Robotics, autonomous vehicles, energy infrastructures, and large-scale cyber-physical systems. By learning control policies directly from interactions with the environment, RL reduces reliance on explicit system modeling and enables adaptive behavior in highly nonlinear and uncertain settings. As a result, Reinforcement Learning has been increasingly adopted in safety-critical domains where classical model-based controllers face scalability limitations and modeling inaccuracies.

Despite these advances, most RL algorithms are built upon the Markov Decision Process (MDP) assumption, which presumes that future system evolution depends solely on the current state and action. In practical networked control systems, however, communication constraints such as sensing delays, actuation latencies, packet losses, and asynchronous information exchange are unavoidable. These temporal imperfections introduce memory effects into the dynamics, effectively violating the Markov property and often leading to degraded performance or even instability when standard RL methods are applied directly. In multi-agent systems (such as vehicular platoons, robotic swarms, or distributed energy networks), delays are further compounded by decentralized communication and heterogeneous network conditions.

\begin{figure}[t]
    \centering
    \subfigure[Classical RL with instantaneous feedback (Markovian).]{
        \includegraphics[width=0.85\linewidth]{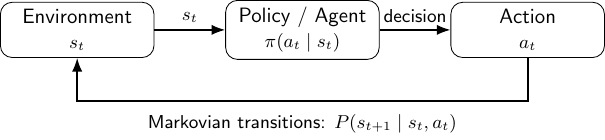}
        \label{subfig:classical}
    }
    \\
    \subfigure[RL under time delays: stale observations and delayed actuation induce non-Markovian dynamics.]{
        \includegraphics[width=\linewidth]{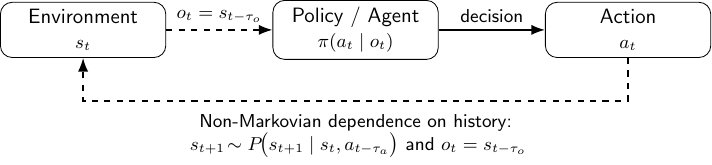}
        \label{subfig:delayed}
    }
    \caption{Conceptual comparison between classical reinforcement learning and reinforcement learning under time delays.}
    \label{fig:rl_delay_comparison}
\end{figure}

To illustrate this fundamental mismatch, Fig.~\ref{subfig:classical} depicts the classical RL interaction loop, in which the agent observes the current system state $s_t$ and applies control actions $a_t$ instantaneously, resulting in Markovian state transitions. In contrast, Fig.~\ref{subfig:delayed} shows a delayed control loop where observations $o_t$ correspond to past system states and actions are executed after non-negligible latencies. These delayed feedback mechanisms introduce an explicit dependence on historical states and actions, thereby transforming the control problem into an intrinsically non-Markovian decision process.

From a control-theoretic perspective, time-delay systems are known to exhibit rich and potentially destabilizing dynamics, where even small latencies may significantly alter stability margins and transient behavior. While classical control theory has developed extensive tools for delay compensation, robust stabilization, and predictor-based feedback, integrating these principles with data-driven RL remains challenging. Naively training RL agents in delayed environments frequently results in slow convergence, oscillatory behavior, and unsafe exploration, underscoring the need for principled delay-aware learning frameworks.

Over the past decade, a growing body of research has sought to extend RL methodologies to address delayed observations, delayed actions, and delayed inter-agent communication. Proposed solutions range from explicit state augmentation and recurrent memory architectures to model-based prediction schemes, robust and domain-randomized training strategies, and safety-constrained hybrid control designs. However, these contributions are dispersed across diverse application domains and often rely on heterogeneous assumptions and evaluation protocols, making it difficult to obtain a unified understanding of their relative strengths, limitations, and practical applicability.

Hence, to the best of our knowledge, this is the first survey that systematically organizes RL methods for control systems affected by time delays across modeling, algorithmic, robustness, and safety perspectives. Its main contributions are summarized as follows:
\begin{itemize}
    \item We introduce a unified modeling perspective for time delays in RL, formally characterizing observation delays, action delays, communication delays, and stochastic latency effects, and clarifying their implications for the Markov property.

    \item We propose a structured taxonomy of delay-aware RL approaches, organizing the literature into five major methodological families: $(i)$ delay-aware state representations, $(ii)$ recurrent memory-based policies, $(iii)$ predictor-based and model-aware learning strategies, $(iv)$ robust and domain-randomized training paradigms, and $(v)$ safe RL frameworks.

    \item We conduct a comparative analysis of existing methods, highlighting key trade-offs in scalability, sample efficiency, robustness, and safety guarantees, and providing practical guidelines for selecting suitable techniques under different delay characteristics and application requirements.

    \item Finally, we identify open challenges and promising research directions, including stability certification, learning under large and stochastic delays, multi-agent coordination under communication constraints, and standardized benchmarking for delay-aware RL.
\end{itemize}

The remainder of this paper is organized as follows. Section~\ref{sec:modeling} introduces the main modeling frameworks for time delays in Reinforcement Learning, formalizing observation, action, and communication delays and their impact on the Markov property. Next, Section~\ref{sec:methods} reviews and categorizes the principal methodological approaches for delay-aware RL. Section~\ref{sec:comparative_analysis} presents a comparative analysis and practical design guidelines. In the sequence, Section~\ref{sec:open_challenges} discusses open challenges and future research directions. And finally, Section~\ref{sec:conclusion} concludes the paper by synthesizing key insights and outlining avenues for future work.

\section{Modeling Time Delays in Reinforcement Learning}
\label{sec:modeling}

Reinforcement Learning is commonly formulated within the framework of Markov Decision Processes (MDPs), defined by the tuple $(\mathcal{S}, \mathcal{A}, P, r, \gamma)$, where $\mathcal{S}$ denotes the state space, $\mathcal{A}$ the action space, $P(s_{t+1}\mid s_t,a_t)$ the transition probability, $r(s_t,a_t)$ the reward function, and $\gamma \in (0,1)$ the discount factor \cite{sutton1998reinforcement}. In this context, the Markov property assumes that the future system evolution depends exclusively on the current state $s_t$ and action $a_t$, enabling algorithms to learn optimal (or good) policies without explicit memory of past interactions \cite{puterman2014markov}. 

In practical cyber-physical and networked control systems, however, information exchange is often affected by sensing delays, actuation latencies, and communication constraints. Such temporal effects introduce dependencies on historical states and actions, rendering the underlying decision process non-Markovian when viewed from the classical RL point-of-view. From a decision-theoretic perspective, delayed feedback constitutes a structured form of partial observability, closely related to Partially Observable Markov Decision Processes (POMDPs), where the agent must infer the current system state from incomplete or outdated information \cite{kaelbling1998planning}. 

Time delays have long been recognized in control theory as a fundamental source of instability and performance degradation. Extensive theoretical tools have been developed to analyze systems with fixed, time-varying, and distributed delays, particularly through Lyapunov-Krasovskii methods and frequency-domain techniques \cite{gu2003stability, richard2003time, fridman2014tutorial, niculescu2002delay}. Even small delays may destabilize otherwise stable systems, highlighting the necessity of delay-aware control strategies. In the following, we categorize the main forms of time delays encountered in RL-based control literature and formalize their impact on the learning process.

\subsection{Observation Delays}

Observation delays arise when the agent receives outdated measurements of the environment state. Let $s_t$ denote the true system state at time $t$. With a constant observation delay $\tau_o$, the agent observes
\begin{equation}
    o_t = s_{t-\tau_o},
    \label{eq:delayed_observation}
\end{equation}
instead of the current state $s_t$. Consequently, the policy is conditioned on stale information, i.e., $a_t = \pi(o_t)$, which may significantly degrade performance in rapidly evolving systems. When $\tau_o$ is time-varying, the observation process becomes stochastic and may further involve packet drops or asynchronous updates across multiple sensors \cite{schenato2008optimal}.

Observation delays can be viewed as a particular form of non-Markovian or partially observable decision process, a perspective explored both in classical RL formulations \cite{whitehead1995reinforcement} and in modern deep RL architectures with memory mechanisms \cite{hausknecht2015deep}.
They can also be seen as a special case of impaired observability, where the agent acts on stale or missing state information. Formal regret analyses in such settings demonstrate that efficient learning is still possible with appropriate modifications to the decision model \cite{chen2024efficient}.

\subsection{Action Delays}

Action delays, also referred to as actuation latencies or dead times, occur when control commands issued by the agent are executed by the system after a delay $\tau_a$. In this case, the system dynamics can then be expressed as
\begin{equation}
    s_{t+1} \sim P\big(s_{t+1} \mid s_t, a_{t-\tau_a}\big),
\end{equation}
where the current state transition depends on past actions. Such delays are prevalent in networked robotics, remote control, and industrial automation, where communication and computation times are non-negligible \cite{farajiparvar2020brief}. Predictor-based compensation techniques from classical control aim to reconstruct the current state from delayed measurements, providing conceptual foundations for learning-based delay mitigation strategies \cite{krstic2009delay}.

Action delays have been explicitly investigated in RL, where delayed actuation induces dependence on past control commands and significantly alters closed-loop dynamics. Both deterministic and stochastic action delays have been shown to degrade performance if not explicitly accounted for \cite{ramstedt2019real, bouteiller2020reinforcement}.

\subsection{State and Communication Delays}

In distributed control scenarios, particularly in multi-agent systems, the evolution of each subsystem may depend explicitly on the delayed states of neighboring agents. Denoting the state of agent $i$ by $s_t^i$, a typical delayed interaction takes the form
\begin{equation}
    s_{t+1}^i \sim P\big(s_{t+1}^i \mid s_t^i, s_{t-\tau_s}^j, a_t^i \big), \quad j \in \mathcal{N}_i,
\end{equation}
where $\mathcal{N}_i$ represents the communication neighborhood of agent $i$ and $\tau_s$ is the communication delay. Such structures are ubiquitous in networked control systems and have been shown to fundamentally alter closed-loop dynamics, often requiring co-design of communication and control strategies \cite{hespanha2007survey}. 

These effects are especially relevant for multi-agent RL under asynchronous information exchange. In such context, communication between agents is often subject to latency and bandwidth constraints, which can significantly affect coordination and performance. Recent works have explicitly modeled such delays in multi-agent environments, incorporating delay-aware decision processes and communication policies that adapt to time-lagged message exchange \cite{chen2020delay, yuan2023dacom}.

\subsection{Stochastic Delays, Jitter, and Packet Loss}

In real-world communication networks, delays are rarely constant; instead, latency, jitter, and packet dropouts often vary stochastically due to shared communication resources, interference, and dynamic network traffic conditions. Such time-varying delays and losses have long been recognized as significant factors affecting the performance and stability of networked control systems (NCSs). In particular, random transmission delays and packet losses can degrade closed‐loop performance or even destabilize a controller designed under ideal latency assumptions \cite{zhang2001stability, hespanha2007survey}. These issues motivate stochastic modeling and control design techniques that explicitly account for uncertainty in timing and communication quality.

In the context of learning-based control, stochastic delays introduce additional challenges, as the sequence of received observations and executed actions depends on unpredictable network behavior. Recent research has begun to study RL methods that cope with the randomness in delays and feedback channels. For example, the framework of random delays extends the delayed decision process by treating action and observation lags as stochastic variables, proposing correction techniques for value estimation in such settings \cite{bouteiller2020reinforcement}. Additionally, methods inspired by random-delay Markov decision processes have been developed to transform certain classes of stochastic delay problems into standard RL formulations that can be addressed by existing algorithms.

Beyond purely RL frameworks, the broader control literature on sequence-based and optimal control over networks subject to stochastic delays and packet loss provides important insights into how random communication effects can be modeled and mitigated. In particular, optimal sequence-based control strategies have been proposed for systems operating over TCP-like networks with random delays and packet dropouts, demonstrating that incorporating knowledge of delay distributions into the control design can improve performance over naive approaches \cite{fischer2013optimal}.

Random delays, as encountered in networked control applications, can be modeled by augmenting the state dynamics and sampling process, allowing classical RL algorithms to be applied after suitable correction mechanisms are introduced \cite{du2022random}.

Together, these lines of work highlight that stochastic delays, jitter, and packet loss should not be treated as negligible perturbations. Instead, they form intrinsic aspects of the environment dynamics that influence both learning and control performance. Addressing these stochastic effects remains an important part of delay-aware RL and motivates the robust and predictor-based methods reviewed in later sections.

\subsection{Implications for Markovianity}

Collectively, the aforementioned delay mechanisms induce dependencies on historical states and actions, violating the Markov assumption inherent in standard RL formulations. The effective system evolution may depend on sequences $\{s_{t-k}, a_{t-k}\}_{k=0}^{K}$, transforming the control problem into a history-dependent or partially observable decision process. 

Several works have explicitly formalized learning and planning under delayed feedback, demonstrating how such effects can be incorporated into augmented state representations or belief-based formulations \cite{walsh2009learning, agarwal2021blind}. Addressing this induced non-Markovianity constitutes the central challenge of Reinforcement Learning under time delays and motivates the methodological developments reviewed in the subsequent sections.

\section{Delay-Aware Reinforcement Learning Methods}
\label{sec:methods}

Building upon the delay modeling frameworks introduced in Section~\ref{sec:modeling}, this section reviews and categorizes the main methodological approaches proposed to address time delays in Reinforcement Learning. Although the specific algorithmic implementations vary across application domains, existing works can be broadly grouped into five complementary families: $(i)$ explicit state augmentation and history-based representations, $(ii)$ recurrent policies with learned memory mechanisms, $(iii)$ predictor-based and model-aware control strategies, $(iv)$ robust and domain-randomized training paradigms, and $(v)$ safe RL frameworks with explicit constraint handling. 

Each family seeks to mitigate the non-Markovian effects induced by delays through distinct principles, ranging from recovering approximate Markovianity to compensating for delayed information or enforcing safety during learning and execution. The following subsections systematically analyze these approaches, highlighting their theoretical foundations, practical advantages, and inherent trade-offs.

\subsection{State Augmentation and History-Based Representations}
\label{subsec:state_augmentation}

One of the most direct strategies to handle time delays in RL is to \emph{augment the state representation} with past observations and control actions, thereby recovering an approximately Markovian process in a lifted space. Delays in observation or control actions break the Markov property presumed in classical MDP formulations, but by augmenting the state with historical components, a delayed system can be treated within standard RL frameworks \cite{nath2021revisiting, chen2021delay}.

In the context of delayed RL, the \emph{Delay-Aware Markov Decision Process} (DA-MDP) formalism proposes that a multi-step delayed MDP can be transformed into a standard MDP with an \emph{augmented state}, where the history of past states and actions becomes part of the current state vector \cite{chen2021delay}. This transformation enables the use of traditional RL algorithms with minimal structural changes to the learning pipeline.

Formally, let $s_t$ denote the true environment state at time $t$, and consider an observation delay of $\tau_o$ steps. An augmented state vector can be defined as
\begin{equation}
    \tilde{s}_t = \big[s_t, s_{t-1}, \ldots, s_{t-\tau_o}, a_{t-1}, \ldots, a_{t-\tau_o}\big],
\end{equation}
where past states and actions are concatenated to form a new state representation $\tilde{s}_t$. Here, the key idea is that the transition dynamics of the delayed environment, when expressed in terms of $\tilde{s}_t$, satisfy the Markov property necessary for RL algorithms such as $Q$-learning or actor–critic methods to operate effectively \cite{nath2021revisiting}.

The \emph{state augmentation approach} is widely used in practice when delays are constant or bounded, and when maintaining a finite memory buffer of past interactions is feasible. In addition to canonical RL benchmarks, this method has been adopted in robotic control problems with delayed sensor feedback and in simulated continuous control environments where communication delays affect observations and actuation timing \cite{agarwal2021blind, wu2024variational, malmir2025diarel}.

While state augmentation recovers Markovian dynamics, it comes at the cost of increased dimensionality of the input space, which can degrade sample efficiency and learning performance as the history length increases. Recent advances seek to mitigate this by leveraging \emph{latent encodings} of history or by learning compact representations that summarize relevant delay information without exhaustive stacking of all past observations \cite{wang2023addressing, xia2024deer}. Such approaches combine the principles of state augmentation with deep representation learning to maintain tractability in high-dimensional control tasks.

Overall, state augmentation and history stacking form the foundational building block for delay-aware RL, bridging the gap between theoretical models of delayed MDPs and practical implementations of deep RL in realistic environments.

\subsection{Recurrent Policies and Memory-Augmented Architectures}
\label{subsec:recurrent_policies}

When time delays or asynchronous information flows break the Markov assumption, an alternative to explicit state augmentation presented in the previous section is to endow the policy and/or value function with \emph{internal memory} capable of summarizing relevant history. Recurrent neural networks (RNNs), particularly Long Short-Term Memory (LSTM) architectures, provide a principled way to compress past observations (and possibly past actions) into a latent state that can serve as a sufficient statistic for decision-making in partially observable or delayed settings \cite{lipton2015critical}.

In recurrent RL, the agent maintains a hidden state $h_t$, updated as
\begin{equation}
    h_t = f_\theta(h_{t-1}, o_t, a_{t-1}),
\end{equation}
where $o_t$ is the current observation given by Eq.~\eqref{eq:delayed_observation}, and $a_{t-1}$ is the previous action. The policy and/or value functions then condition on $h_t$, e.g.,
\begin{equation}
    a_t \sim \pi_\theta(\cdot \mid h_t), \qquad V_\theta(h_t),
\end{equation}
so that history-dependent dynamics induced by delays are handled implicitly through memory.

Early deep RL work demonstrated that recurrent architectures can mitigate partial observability by integrating information over time. Deep Recurrent $Q$-Networks (DRQN) replace part of the feedforward $Q$-network with an LSTM to cope with missing or flickering observations, showing that recurrence can serve as an alternative to explicit frame stacking \cite{hausknecht2015deep}. 

In continuous control, the authors of \cite{heess1512memory} proposed memory-based control with recurrent policies by extending deterministic policy gradient and stochastic value gradient methods using recurrent networks trained through backpropagation through time, addressing partially observed control problems that require memory.

Recurrent policies are also widely used in modern actor-critic implementations. For instance, asynchronous actor-critic methods incorporate recurrent components (e.g., A3C LSTMs) to handle partial observability and temporal dependencies in high-dimensional domains \cite{mnih2016asynchronous}. More recently, recurrent off-policy baselines have been systematically benchmarked for memory-based continuous control, implementing recurrent variants of DDPG \cite{lillicrap2015continuous}, TD3 \cite{fujimoto2018addressing} and SAC \cite{haarnoja2018soft} and investigating design choices such as sequence length, burn-in, and replay-buffer handling \cite{yang2021recurrent}.

Compared with explicit history stacking, recurrent architectures can learn \emph{adaptive} temporal representations and may reduce the need to hand-pick a history horizon. 
However, training recurrent policies in RL introduces several additional challenges compared to feedforward architectures. Sequence-based optimization requires storing and processing temporally ordered trajectories, which substantially increases memory usage and computational cost, particularly for long effective delay horizons. Moreover, credit assignment becomes more difficult as relevant rewards may depend on events far in the past, leading to vanishing or noisy gradient signals when backpropagating through extended temporal dependencies \cite{bengio1994learning}. 
In off-policy settings, stability can be highly sensitive to replay-buffer composition, sequence truncation lengths, and hidden-state initialization strategies, as mismatches between training and deployment dynamics may induce biased gradient estimates and slow convergence \cite{yang2021recurrent}.

In practice, recurrent policies are often combined with complementary mechanisms (e.g., explicit buffering, learned predictive state representations, or world models) when delays are large or highly variable \cite{karamzade2024reinforcement}.

\subsection{Predictor-Based and Model-Aware Reinforcement Learning Methods}
\label{subsec:predictor_model}

A complementary class of approaches for handling delays in RL explicitly incorporates a model of the environment or an auxiliary predictive mechanism to estimate future states or compensate for outdated information. Unlike implicit memory mechanisms such as recurrent policies or explicit history stacking, described in the previous sections, predictor-based methods aim to directly forecast the latent state at decision time, effectively mitigating the adverse effects of observation $\tau_o$ or action $\tau_a$ delays.

One salient direction leverages \emph{learned world models} that approximate environment dynamics to predict the current or future state based on delayed and partial observations. For instance, in \cite{karamzade2024reinforcement}, the authors propose using world models to reduce delayed POMDPs to more tractable representations and thereby enable delay-aware RL policy learning. Their methods demonstrated improved performance in continuous control benchmarks with delayed visual observations by explicitly learning dynamics that account for observation latency.

Model-aware RL approaches jointly learn a dynamics model that assimilates the effects of delays into the state transition function. In the context of continuous control tasks with action delays, \cite{chen2021delay} developed a delay-aware model-based RL framework in which the multi-step delay is incorporated into the learned predictive model. The resulting algorithm, Delay-Aware Trajectory Sampling (DATS), explicitly separates known delay effects from unknown environment dynamics, enabling efficient planning and policy optimization despite time lags.

Predictor-based strategies can also be combined with filtering techniques that maintain estimated latent states in the presence of random or variable delays. Recent work on model-based RL under random observation delays formulates a filtering process that updates the agent’s belief state from out-of-sequence or stochastically arriving observations and integrates this process into the learning loop, leading to enhanced resilience to delay variability compared with naive stacking methods \cite{karamzade2025model}.

These predictor and model-aware methods share a central idea: by shifting the learning problem from raw interaction to \emph{informed prediction}, delayed observations and actions can be compensated for before policy evaluation and improvement \cite{wu2025directly}. Such methods are particularly advantageous when delays are significant and structured (e.g., consistent latencies in sensor feedback or action execution) because an accurate predictive model can reduce the effective feedback lag perceived by the agent.

However, model learning and prediction incur their own challenges. Accurate dynamics estimation can be computationally expensive, especially in high-dimensional continuous control domains, and model errors can propagate into policy updates if not properly regularized. Research in this area thus often combines model learning with uncertainty estimation, planning, or conservative updates to balance prediction accuracy with stable policy improvement \cite{luo2024survey}.

Recent advances in neural world models have demonstrated that accurate multi-step prediction can significantly improve control performance in high-dimensional continuous domains \cite{janner2019trust,hafner2019dream}. Such predictive representations are particularly well suited for delay-affected systems, as delayed observations can be propagated forward in time to reconstruct approximate current states prior to action selection \cite{agarwal2021blind, chen2024efficient}.

Overall, predictor-based and model-aware RL techniques constitute a powerful class of solutions when the environment exhibits predictable delay patterns that can be approximated by a learned or analytic dynamics model.

\subsection{Robust and Domain-Randomized Training}
\label{subsec:robust}

Robust training strategies aim to produce policies that maintain satisfactory performance under variations in environment dynamics, sensing conditions, and external disturbances \cite{huang2022robust,neto2024robust}. Foundational surveys characterize robust RL approaches by their treatment of transition, disturbance, action, and observation uncertainty, and relate them to concepts such as distributional robustness and risk-sensitive optimization \cite{moos2022robust}. In systems affected by communication delays, jitter, and packet loss, robustness becomes particularly important, as the temporal characteristics of information flow may vary across operating conditions and network loads.

A widely adopted robustness technique is \emph{domain randomization}, originally introduced in robotics to bridge the gap between simulation and real-world deployment \cite{miranda2023generalization}. In this paradigm, key environment parameters are randomly sampled during training so that the agent experiences a diverse distribution of dynamics and perturbations. The resulting policy is, therefore, optimized over a family of environments rather than a single nominal model, improving generalization to unseen conditions \cite{slaoui2019robust,zhao2021robust,ajani2023evaluating}.

In the context of delayed RL, domain randomization can be naturally extended to temporal parameters by sampling delay magnitudes, jitter distributions, or packet-drop probabilities during training. Let $\phi$ denote a vector of environment parameters including delay characteristics, and $p(\phi)$ their training distribution. The robust policy optimization problem can be expressed as
\begin{equation}
    \max_{\theta} \; \mathbb{E}_{\phi \sim p(\phi)} \left[ \mathbb{E}_{\tau \sim \pi_\theta} \left[ R(\tau; \phi) \right] \right],
\end{equation}
where the policy $\pi_\theta$ is optimized to perform well across the full range of delay realizations. By repeatedly exposing the agent to diverse temporal behaviors, domain randomization encourages the emergence of delay-resilient control strategies without explicit delay modeling.

Beyond randomization, robust RL frameworks explicitly optimize policies under worst-case or adversarial perturbations. Early work introduced robustness by incorporating uncertainty directly into the learning objective, enabling agents to handle model mismatches and disturbances \cite{morimoto2005robust}. More recent approaches train policies against adversarial dynamics or populations that actively seek to degrade performance, thereby improving resilience to extreme variations and unmodeled effects \cite{vinitsky2020robust, pattanaik2017robust}.

Although most robust RL formulations were not originally developed with communication delays in mind, delays can be naturally interpreted as structured uncertainty in state transitions or observation processes. Consequently, robust training techniques provide a complementary tool for mitigating the impact of stochastic or variable latency in networked control systems \cite{wang2021online}.

The main advantage of robust and domain-randomized training lies in its simplicity and compatibility with standard deep RL algorithms such as PPO \cite{schulman2017proximal}, SAC and TD3. These strategies require no architectural changes to incorporate memory or predictors, relying instead on exposure to diverse training conditions to induce robustness. However, they typically increase sample complexity and require careful design of the randomization or adversarial perturbation distributions to ensure coverage of realistic delay scenarios.

Other works in this area include analysis of sample complexity for robust policy learning with generative models \cite{panaganti2022sample}. Recent extensions also incorporate robustness into MARL, where agents must contend with uncertainty in both environment dynamics and other agents’ policies \cite{zhang2020robust}. These robust formulations naturally complement delay-aware RL, as communication latency and stochastic timing can be interpreted as structured model uncertainty.

Overall, robust and domain-randomized training offers a practical and scalable pathway to enhance the reliability of RL controllers operating under uncertain and time-varying delay conditions.

\subsection{Safe Reinforcement Learning under Time Delays}
\label{subsec:safe_rl}

Safety constraints are central in delay-affected control systems, as outdated measurements or delayed actuation can induce overshoots, constraint violations, and instability, even when nominal performance is acceptable. 
In Safe Reinforcement Learning (Safe-RL), the goal is to learn policies that maximize return while respecting safety constraints during training and/or deployment \cite{garcia2015comprehensive}. A generic formulation takes the form
\begin{align}
    \max_{\pi} \quad & \mathbb{E}_\pi \Big[\sum_{t=0}^{\infty} \gamma^t r(s_t,a_t)\Big],\\
    \text{s.t.} \quad & \mathbb{E}_\pi \Big[\sum_{t=0}^{\infty} \gamma^t c_i(s_t,a_t)\Big] \le d_i, ~i=1,\dots,m, \nonumber
\end{align}
where $r(\cdot)$ denotes the performance reward, $c_i(\cdot)$ represents safety-related cost signals, and $d_i$ specifies admissible risk levels. This framework naturally captures delayed safety effects, since violations caused by latency or outdated information are accumulated over time rather than enforced instantaneously.

A classical perspective organizes Safe-RL methods into approaches that modify the learning objective and approaches that modify the exploration/execution process.
A common formalism for Safe-RL is the Constrained Markov Decision Process (CMDP), which augments the reward objective with constraints on expected cumulative costs \cite{kushwaha2025survey}. Policy optimization under CMDPs can be addressed via trust-region methods with constraint satisfaction guarantees, such as Constrained Policy Optimization (CPO) \cite{achiam2017constrained}. While CMDP formulations are not specific to delays, they naturally accommodate delay-induced risk by defining constraint costs that capture safety violations (e.g., inter-vehicle distance constraints in platoons) under delayed observations/actions.

Beyond trust-region CMDP methods, Lyapunov-based Safe-RL provides a constructive way to ensure constraint satisfaction by building a Lyapunov function and enforcing local constraints that imply global safety. In \cite{chow2018lyapunov}, the authors propose a Lyapunov-based approach that transforms dynamic programming and RL algorithms into safe counterparts through linear constraints derived from Lyapunov conditions. In delay-prone settings, Lyapunov-style reasoning is appealing because time delays can be interpreted as uncertainty or unmodeled dynamics that reduce stability margins. In other words, Lyapunov constraints can be used to restrict policy updates to remain within a safe region.

Another practical family of methods enforces safety by \emph{filtering} the RL action through an optimization layer that minimally modifies the proposed action to satisfy constraints. A representative example is the \emph{safety layer} approach in \cite{dalal2018safe}, which projects actions onto a constraint-satisfying set using a per-step correction based on a learned local model of constraint dynamics. Similarly, \emph{OptLayer} in \cite{pham2018optlayer} embeds a constrained optimization module into the policy pipeline in robotic manipulation scenarios, outputting the closest feasible action to the network prediction. These action-correction layers are particularly useful under delays, since delayed information can cause transient violations; an online correction step can act as a last-resort guardrail.

On the other hand, Control Barrier Functions (CBFs) provide a control-theoretic framework to enforce forward invariance of safe sets via real-time safety filters, often implemented as quadratic programs that minimally modify a nominal controller \cite{ames2019control}. Learning-based controllers combined with CBF filters have become a prominent direction for safety-critical autonomy. For example, in \cite{taylor2020learning}, the authors discuss learning for safety-critical control with CBFs and related formulations. In delayed or sampled-data implementations, CBF-based filters can be adapted to account for discretization and information staleness, thereby improving safety under communication constraints.

In addition to instantaneous safety filters such as CBFs, a large body of the Safe-RL literature formulates safety requirements through CMDPs, where cumulative cost signals represent violations of safety or resource limits. These problems are commonly addressed via Lagrangian relaxation techniques that transform constrained optimization into a dual learning problem, enabling policy updates that balance reward maximization and constraint satisfaction \cite{achiam2017constrained, ray2019benchmarking}. In delayed control systems, where safety violations may manifest after significant temporal offsets due to communication or actuation latency, such cumulative constraint formulations provide a natural mechanism to absorb delayed feedback and enforce long-term safety guarantees.
More recently, control-theoretic approaches based on control barrier functions have gained prominence as real-time safety filters that minimally modify learned policies to maintain forward invariance of safe sets \cite{emam2022safe}. 

In summary, Safe-RL methods provide complementary tools to delay-aware representations and robust training: CMDP/Lagrangian and Lyapunov methods incorporate safety into the learning objective and updates, while safety layers and CBF filters enforce constraints at execution time. For delay-affected multi-agent control, hybrid designs that combine delay-aware state representations with a safety filter (backup controller or CBF-QP) are often an effective practical compromise, providing both performance learning and explicit safety guarantees.

\section{Comparative Analysis and Practical Guidelines}
\label{sec:comparative_analysis}

The methods reviewed in the previous section address time delays in Reinforcement Learning from different perspectives, ranging from explicit state representations to memory-based architectures, predictive modeling, robustness-oriented training, and safety-constrained control. Each approach offers distinct advantages and limitations depending on the magnitude, variability, and structure of delays, as well as on system complexity and safety requirements. This section provides a comparative analysis of the main method families and offers practical guidelines for their application.

\subsection{Conceptual Comparison of Frameworks and Trade-Offs}

While each class of approaches seeks to mitigate the detrimental effects of delayed observations, actions, or communication, they differ significantly in terms of modeling assumptions, computational complexity, stability properties, data efficiency, and ease of deployment in real-world systems. A conceptual comparison is therefore essential to clarify the strengths and limitations of each strategy and to guide practitioners in selecting appropriate techniques for specific delay characteristics and application domains. Here, we systematically contrast all frameworks along key dimensions such as delay structure, scalability, training stability, and robustness to uncertainty.

State augmentation techniques discussed in Section~\ref{subsec:state_augmentation} explicitly encode historical information into the policy input, effectively recovering approximate Markovianity at the cost of increased state dimensionality. These methods are simple to implement and compatible with standard deep RL algorithms, but can suffer from reduced sample efficiency as the memory horizon grows.

Recurrent policies presented in Section~\ref{subsec:recurrent_policies} provide a learned, compact representation of history, allowing the agent to adaptively extract relevant temporal dependencies. While often more parameter-efficient than explicit stacking, recurrent architectures introduce training complexity and sensitivity to sequence length, replay-buffer handling, and optimization stability.

Predictor-based and model-aware approaches in Section~\ref{subsec:predictor_model} seek to compensate for delays by forecasting latent system states or embedding delay dynamics into learned models. These methods can significantly reduce the effective feedback lag and improve performance under structured delays, but rely on accurate model learning and may degrade when prediction errors accumulate.

Meanwhile, robust and domain-randomized training in Section~\ref{subsec:robust} exposes policies to a distribution of delay conditions during training, encouraging generalization across varying temporal behaviors. This strategy is highly scalable and easy to integrate with existing RL algorithms, yet typically requires increased sample complexity and careful design of randomization distributions.

Finally, Safe-RL methods in Section~\ref{subsec:safe_rl} introduce explicit constraint handling mechanisms, ensuring stability and safety despite delay-induced uncertainty. These approaches are essential for safety-critical systems but may restrict exploration and reduce achievable performance when constraints are conservative.


{\renewcommand{\arraystretch}{1.5}
\begin{table*}[t]
\centering
\caption{Qualitative comparison of main RL approaches for time-delay systems.}
\label{tab:delay_comparison}
\resizebox{\linewidth}{!}{
\begin{tabular}{C{2.5cm} C{2.0cm} C{2.0cm} C{2.0cm} C{2.0cm} C{2.0cm}}
\toprule
\bf Method 
& \bf Markov Recovery 
& \bf Scalability 
& \bf Sample Efficiency 
& \bf Safety Guarantees 
& \bf Implemen-tation Effort \\
\midrule
State Augmentation 
& High 
& Medium-Low 
& Medium-Low 
& None 
& Low \\
\hline
Recurrent Policies 
& High 
& High 
& Medium 
& None 
& Medium \\
\hline
Predictor-Based RL 
& High 
& Medium 
& High (model dependent) 
& None 
& High \\
\hline
Domain Randomization 
& Partial 
& High 
& Low 
& None 
& Low \\
\hline
Safe-RL
& High 
& Medium 
& Medium 
& Strong 
& High \\
\bottomrule
\end{tabular}
}
\end{table*}
}

To synthesize the diverse body of delay-aware RL methods discussed in Section~\ref{sec:methods}, Table~\ref{tab:delay_comparison} provides a qualitative comparison across key design dimensions, including Markovian recovery capability, scalability, sample efficiency, safety guarantees, and implementation effort. Rather than ranking approaches by absolute performance, the table highlights structural trade-offs that consistently emerge across application domains.

The table shows that explicit state augmentation offers a straightforward path to recover approximate Markovianity, but its scalability is limited by the rapid growth of input dimensionality as the history horizon increases. Recurrent policies alleviate this dimensionality burden by learning compact memory representations, thereby scaling more favorably to high-dimensional and multi-agent systems, albeit with increased training complexity.

Predictor-based and model-aware approaches tend to achieve high sample efficiency when accurate dynamics models can be learned or specified, as delay compensation reduces the effective feedback lag perceived by the agent. However, their reliance on model fidelity introduces sensitivity to prediction errors, particularly in highly nonlinear or stochastic environments.

Robust and domain-randomized training methods, while not explicitly recovering Markovian structure, demonstrate strong generalization to variable delay conditions by optimizing performance across a distribution of temporal perturbations. This robustness often comes at the expense of higher sample complexity, as the agent must explore a broader range of environment realizations during training.

Finally, Safe-RL frameworks stand out in their ability to provide formal or empirical safety guarantees, making them indispensable for safety-critical applications such as autonomous driving and robotic swarms. Nevertheless, conservative constraint enforcement may limit achievable performance and increase computational overhead.

Overall, Table~\ref{tab:delay_comparison} illustrates that no single approach dominates across all criteria. Instead, practical delay-aware RL systems often benefit from hybrid designs that combine complementary strengths (recurrent policies trained with domain randomization and safeguarded by safety filters, for example).

\subsection{Practical Guidelines for Delay-Aware RL Design}

Based on the surveyed literature, the following guidelines can assist practitioners and researchers in selecting appropriate methods:
\begin{itemize}
    \item \emph{Small and fixed delays:} When delays are short and known a priori, state augmentation or simple recurrent policies are typically sufficient and easy to deploy. In such cases, incorporating a finite history of past states and actions increases the effective state dimension only linearly with the delay horizon, preserving tractability while restoring approximate Markovianity. Moreover, fixed delays enable deterministic buffering strategies and stable feedforward policy learning without requiring complex prediction or robustness mechanisms.

    \item \emph{Large but structured delays:} When delays span long horizons but follow consistent or predictable patterns, predictor-based and model-aware methods become particularly effective. By explicitly learning system dynamics and propagating delayed measurements forward in time, these approaches compensate for feedback latency and reconstruct approximate current states before policy evaluation. This reduces the effective non-Markovianity perceived by the agent and often yields substantially improved performance compared to purely memory-based strategies.
    
    \item \emph{Variable or stochastic delays:} When delays are unpredictable, explicit modeling becomes unreliable, and robustness is best achieved through domain randomization and memory-based policies that adapt across varying latency conditions.

    \item \emph{Safety-critical systems:} In applications where instability or constraint violations can lead to physical damage or unacceptable risk, explicit safety mechanisms are essential when learning under time delays. Safe-RL frameworks based on constrained optimization, as well as hybrid control designs incorporating real-time safety filters such as control barrier functions, should be employed to enforce safety guarantees while allowing policy improvement. These approaches prevent catastrophic behaviors induced by delayed feedback and ensure reliable operation throughout both training and deployment.

    \item \emph{Large-scale multi-agent systems:} In networked environments with many interacting agents, explicitly augmenting the state with delayed histories quickly leads to prohibitively high-dimensional representations that scale with the number of agents, state dimension, and delay horizon. Recurrent policy architectures, which compactly encode temporal information through shared memory, together with robust training strategies that tolerate variability in communication timing, offer more scalable and practical solutions for handling delays in large multi-agent systems.

    \item \emph{Real-time and computational constraints:} In embedded or fast control loops, the computational latency of inference and auxiliary modules must remain significantly below the system sampling period. Lightweight feedforward policies and bounded-history augmentation are generally preferable for strict real-time requirements, whereas predictor-based models and recurrent architectures should be carefully optimized to avoid excessive rollout or memory overhead. Practical deployment, therefore, requires balancing delay compensation accuracy with real-time feasibility.
\end{itemize}

Hybrid approaches that combine multiple strategies (for instance, delay-aware state representations with safety filters or robust training) have shown particular promise in complex real-world applications and represent a growing research direction.

\section{Open Challenges and Future Research Directions}
\label{sec:open_challenges}

Despite significant progress in Delay-aware Reinforcement Learning, several theoretical and practical challenges remain open. Addressing these issues is essential for the reliable deployment of RL-based controllers in large-scale, safety-critical, and networked systems subject to communication latency and temporal uncertainty.

\subsection{Formal Stability and Performance Guarantees}

Most existing delay-aware RL methods rely on empirical performance improvements without providing formal guarantees on closed-loop stability or convergence rates. While classical control theory offers rich tools for analyzing time-delay systems, integrating these guarantees with learning-based controllers remains largely unresolved. 

Promising directions include Lyapunov-guided policy learning, certified RL frameworks, and hybrid designs that blend learned policies with provably stable backup controllers. Developing scalable methods that provide stability certificates under variable and stochastic delays is a critical research frontier.

\subsection{Learning under Large and Highly Variable Delays}

Current techniques perform well for moderate delays but often degrade when latency becomes large or highly stochastic. Recurrent architectures and predictor-based models struggle with long memory horizons and compounding prediction errors, while state augmentation becomes impractical due to dimensionality growth. Future research should explore compact latent representations, hierarchical memory structures, and delay-aware belief-state formulations capable of handling long-range temporal dependencies efficiently.

\subsection{Multi-Agent Systems and Communication Constraints}

In distributed systems such as vehicle platoons, robotic swarms, and sensor networks, delays interact with network topology, bandwidth limitations, and asynchronous updates \cite{foerster2016learning,zhang2021multi}. Most existing RL frameworks assume simplified or fixed communication patterns, limiting scalability and realism. Open challenges include co-designing communication protocols and control policies, learning under partial and intermittent connectivity, and ensuring collective stability when information arrives with heterogeneous delays across agents.

\subsection{Safety under Delayed and Uncertain Feedback}

While Safe-RL methods provide mechanisms for constraint enforcement, their integration with delay-aware learning remains incomplete. Delays can undermine safety guarantees by introducing outdated state information into safety filters and constraint estimators. Developing delay-robust safety layers, sampled-data barrier functions, and real-time certification mechanisms that explicitly account for information staleness is crucial for real-world deployment.

\subsection{Benchmarking and Standardized Evaluation}

A major limitation in the current literature is the absence of standardized benchmarking frameworks explicitly designed for Reinforcement Learning under time delays. Most existing studies rely on custom-built environments with task-specific latency models, which hampers systematic comparison across methods and compromises reproducibility. The development of open-source benchmark suites that systematically vary delay magnitude, jitter, and packet loss—across both single-agent and multi-agent control domains—would significantly facilitate fair evaluation and accelerate progress in delay-aware reinforcement learning research.

Despite the absence of unified benchmarking frameworks, several experimental testbeds and modified control environments have been proposed to evaluate RL under delayed feedback. Common approaches include delayed variants of classical continuous-control tasks, such as inverted pendulum, cart-pole \cite{towers2024gymnasium}, and MuJoCo locomotion environments \cite{todorov2012mujoco}, where observation and action buffers emulate fixed or stochastic communication latency. More recent studies(such as \cite{wang2023addressing}\footnote{\url{https://github.com/microsoft/Addressing-signal-delay-in-deep-RL}} and \cite{chen2020delay}\footnote{\url{https://github.com/baimingc/delay-aware-MAR}}) have also explored networked multi-agent scenarios, including vehicle platoons and robotic swarms, with explicit modeling of transmission delays and packet loss. While these experimental setups provide valuable insights into algorithm behavior under latency, their heterogeneous delay models, performance metrics, and task definitions limit direct comparability across studies.

\subsection{Sim-to-Real Transfer with Temporal Uncertainty}

Transferring delay-aware RL policies from simulation to physical systems remains challenging due to discrepancies in communication latency, sensor update rates, and hardware timing \cite{tobin2017domain,chebotar2019closing}. While domain randomization partially addresses these gaps, more principled approaches that explicitly model and adapt to real-time delay profiles are needed. Online delay identification and adaptive policy modulation represent promising research directions for closing the sim-to-real temporal gap.

Overall, advancing delay-aware RL requires tighter integration between control theory, communication systems, and modern RL methodologies. Progress along these directions will be essential for deploying learning-based controllers in complex, time-sensitive cyber-physical systems.

\section{Conclusion}
\label{sec:conclusion}

This survey presented a comprehensive review of Reinforcement Learning methods for control systems affected by time delays. By formalizing observation, action, and communication delays and analyzing their impact on the Markov property, we established a unified modeling perspective for delay-affected RL problems. Existing approaches were systematically categorized into five methodological families: state augmentation, recurrent policies, predictor-based and model-aware methods, robust and domain-randomized training strategies, and Safe-RL frameworks.

Through a comparative analysis, we highlighted the fundamental trade-offs among these approaches in terms of scalability, sample efficiency, robustness, and safety guarantees, and provided practical guidelines for selecting suitable techniques under different delay characteristics and application requirements. The discussion of open challenges emphasized the need for tighter integration between Reinforcement Learning, control theory, and communication systems, particularly with respect to formal stability certification, large-delay learning, multi-agent coordination under communication constraints, and standardized benchmarking.

Overall, Delay-aware RL remains a rapidly evolving research area with strong potential for real-world impact across autonomous systems, Robotics, and cyber-physical infrastructures. We hope that this survey serves as a unified reference and a catalyst for future research toward reliable, safe, and scalable learning-based control in the presence of time delays.

\section*{Acknowledgment}
 
This work was supported in part by the Coordenação de Aperfeiçoamento de Pessoal de Nível Superior (CAPES), Brazil (Finance Code 001), the Conselho Nacional de Desenvolvimento Científico e Tecnológico (CNPq), Brazil, and the Fundação de Amparo à Pesquisa do Estado de Minas Gerais (FAPEMIG), Brazil.


\bibliographystyle{elsarticle-num} 
\bibliography{07_references}

@book{sutton1998reinforcement,
  title={Reinforcement learning: An introduction},
  author={Sutton, Richard S and Barto, Andrew G and others},
  volume={1},
  number={1},
  year={1998},
  publisher={MIT press Cambridge}
}

@book{puterman2014markov,
  title={Markov decision processes: discrete stochastic dynamic programming},
  author={Puterman, Martin L},
  year={2014},
  publisher={John Wiley \& Sons}
}

@article{kaelbling1998planning,
  title={Planning and acting in partially observable stochastic domains},
  author={Kaelbling, Leslie Pack and Littman, Michael L and Cassandra, Anthony R},
  journal={Artificial intelligence},
  volume={101},
  number={1-2},
  pages={99--134},
  year={1998},
  publisher={Elsevier}
}

@book{gu2003stability,
  title={Stability of time-delay systems},
  author={Gu, Keqin and Chen, Jie and Kharitonov, Vladimir L},
  year={2003},
  publisher={Springer Science \& Business Media}
}

@article{richard2003time,
  title={Time-delay systems: an overview of some recent advances and open problems},
  author={Richard, Jean-Pierre},
  journal={Automatica},
  volume={39},
  number={10},
  pages={1667--1694},
  year={2003},
  publisher={Elsevier}
}

@article{fridman2014tutorial,
  title={Tutorial on Lyapunov-based methods for time-delay systems},
  author={Fridman, Emilia},
  journal={European Journal of Control},
  volume={20},
  number={6},
  pages={271--283},
  year={2014},
  publisher={Elsevier}
}

@book{niculescu2002delay,
  title={Delay effects on stability: a robust control approach},
  author={Niculescu, Silviu-Iulian},
  year={2002},
  publisher={Springer}
}

@book{krstic2009delay,
  title={Delay compensation for nonlinear, adaptive, and PDE systems},
  author={Krstic, Miroslav},
  year={2009},
  publisher={Springer}
}

@article{hespanha2007survey,
  title={A survey of recent results in networked control systems},
  author={Hespanha, Joo P and Naghshtabrizi, Payam and Xu, Yonggang},
  journal={Proc. of the IEEE},
  volume={95},
  number={1},
  pages={138--162},
  year={2007},
  publisher={IEEE}
}

@article{walsh2009learning,
  title={Learning and planning in environments with delayed feedback},
  author={Walsh, Thomas J and Nouri, Ali and Li, Lihong and Littman, Michael L},
  journal={Autonomous Agents and Multi-Agent Systems},
  volume={18},
  number={1},
  pages={83--105},
  year={2009},
  publisher={Springer}
}

@article{agarwal2021blind,
  title={Blind decision making: Reinforcement learning with delayed observations},
  author={Agarwal, Mridul and Aggarwal, Vaneet},
  journal={Pattern Recognition Letters},
  volume={150},
  pages={176--182},
  year={2021},
  publisher={Elsevier}
}

@article{whitehead1995reinforcement,
  title={Reinforcement learning of non-Markov decision processes},
  author={Whitehead, Steven D and Lin, Long-Ji},
  journal={Artificial Intelligence},
  volume={73},
  number={1-2},
  pages={271--306},
  year={1995},
  publisher={Elsevier}
}

@inproceedings{hausknecht2015deep,
  title={Deep Recurrent Q-Learning for Partially Observable MDPs},
  author={Hausknecht, Matthew J and Stone, Peter},
  booktitle={AAAI fall symposia},
  volume={45},
  pages={141},
  year={2015}
}

@article{ramstedt2019real,
  title={Real-time reinforcement learning},
  author={Ramstedt, Simon and Pal, Chris},
  journal={Advances in neural information processing systems},
  volume={32},
  year={2019}
}

@inproceedings{bouteiller2020reinforcement,
  title={Reinforcement learning with random delays},
  author={Bouteiller, Yann and Ramstedt, Simon and Beltrame, Giovanni and Pal, Christopher and Binas, Jonathan},
  booktitle={International conference on learning representations},
  year={2020}
}

@article{chen2020delay,
  title={Delay-aware multi-agent reinforcement learning for cooperative and competitive environments},
  author={Chen, Baiming and Xu, Mengdi and Liu, Zuxin and Li, Liang and Zhao, Ding},
  journal={arXiv preprint arXiv:2005.05441},
  year={2020}
}

@inproceedings{yuan2023dacom,
  title={DACOM: Learning delay-aware communication for multi-agent reinforcement learning},
  author={Yuan, Tingting and Chung, Hwei-Ming and Yuan, Jie and Fu, Xiaoming},
  booktitle={AAAI Conference on Artificial Intelligence},
  volume={37},
  number={10},
  pages={11763--11771},
  year={2023}
}

@article{zhang2001stability,
  title={Stability of networked control systems},
  author={Zhang, Wei and Branicky, Michael S and Phillips, Stephen M},
  journal={IEEE control systems magazine},
  volume={21},
  number={1},
  pages={84--99},
  year={2001},
  publisher={IEEE}
}

@inproceedings{fischer2013optimal,
  title={Optimal sequence-based LQG control over TCP-like networks subject to random transmission delays and packet losses},
  author={Fischer, J{\"o}rg and Hekler, Achim and Dolgov, Maxim and Hanebeck, Uwe D},
  booktitle={American Control Conference},
  pages={1543--1549},
  year={2013},
  organization={IEEE}
}

@article{schenato2008optimal,
  title={Optimal estimation in networked control systems subject to random delay and packet drop},
  author={Schenato, Luca},
  journal={IEEE transactions on automatic control},
  volume={53},
  number={5},
  pages={1311--1317},
  year={2008},
  publisher={IEEE}
}

@article{farajiparvar2020brief,
  title={A brief survey of telerobotic time delay mitigation},
  author={Farajiparvar, Parinaz and Ying, Hao and Pandya, Abhilash},
  journal={Frontiers in Robotics and AI},
  volume={7},
  pages={578805},
  year={2020},
  publisher={Frontiers Media SA}
}

@inproceedings{nath2021revisiting,
  title={Revisiting state augmentation methods for reinforcement learning with stochastic delays},
  author={Nath, Somjit and Baranwal, Mayank and Khadilkar, Harshad},
  booktitle={ACM international conference on information \& knowledge management},
  pages={1346--1355},
  year={2021}
}

@article{chen2021delay,
  title={Delay-aware model-based reinforcement learning for continuous control},
  author={Chen, Baiming and Xu, Mengdi and Li, Liang and Zhao, Ding},
  journal={Neurocomputing},
  volume={450},
  pages={119--128},
  year={2021},
  publisher={Elsevier}
}

@article{wu2024variational,
  title={Variational delayed policy optimization},
  author={Wu, Qingyuan and Zhan, Simon S and Wang, Yixuan and Wang, Yuhui and Lin, Chung-Wei and Lv, Chen and Zhu, Qi and Huang, Chao},
  journal={Advances in neural information processing systems},
  volume={37},
  pages={54330--54356},
  year={2024}
}

@article{xia2024deer,
  title={Deer: A delay-resilient framework for reinforcement learning with variable delays},
  author={Xia, Bo and Kong, Yilun and Chang, Yongzhe and Yuan, Bo and Li, Zhiheng and Wang, Xueqian and Liang, Bin},
  journal={arXiv preprint arXiv:2406.03102},
  year={2024}
}

@inproceedings{wang2023addressing,
  title={Addressing signal delay in deep reinforcement learning},
  author={Wang, Wei and Han, Dongqi and Luo, Xufang and Li, Dongsheng},
  booktitle={International Conference on Learning Representations},
  year={2023}
}

@article{malmir2025diarel,
  title={DiAReL: Reinforcement Learning with Disturbance Awareness for Robust Sim2Real Policy Transfer in Robot Control},
  author={Malmir, Mohammadhossein and Josifovski, Josip and Klarmann, Noah and Knoll, Alois},
  journal={IEEE Transactions on Control Systems Technology},
  year={2025},
  publisher={IEEE}
}

@article{lipton2015critical,
  title={A critical review of recurrent neural networks for sequence learning},
  author={Lipton, Zachary C and Berkowitz, John and Elkan, Charles},
  journal={arXiv preprint arXiv:1506.00019},
  year={2015}
}

@article{wang2021online,
  title={Online robust reinforcement learning with model uncertainty},
  author={Wang, Yue and Zou, Shaofeng},
  journal={Advances in Neural Information Processing Systems},
  volume={34},
  pages={7193--7206},
  year={2021}
}

@article{schulman2017proximal,
  title={Proximal policy optimization algorithms},
  author={Schulman, John and Wolski, Filip and Dhariwal, Prafulla and Radford, Alec and Klimov, Oleg},
  journal={arXiv preprint arXiv:1707.06347},
  year={2017}
}

@article{heess1512memory,
  title={Memory-based control with recurrent neural networks},
  author={Heess, Nicolas and Hunt, Jonathan J and Lillicrap, Timothy P and Silver, David},
  journal={arXiv preprint arXiv:1512.04455},
  year={2015}
}

@inproceedings{mnih2016asynchronous,
  title={Asynchronous methods for deep reinforcement learning},
  author={Mnih, Volodymyr and Badia, Adria Puigdomenech and Mirza, Mehdi and Graves, Alex and Lillicrap, Timothy and Harley, Tim and Silver, David and Kavukcuoglu, Koray},
  booktitle={International conference on machine learning},
  pages={1928--1937},
  year={2016},
  organization={PmLR}
}

@article{yang2021recurrent,
  title={Recurrent off-policy baselines for memory-based continuous control},
  author={Yang, Zhihan and Nguyen, Hai},
  journal={arXiv preprint arXiv:2110.12628},
  year={2021}
}

@article{karamzade2024reinforcement,
  title={Reinforcement learning from delayed observations via world models},
  author={Karamzade, Armin and Kim, Kyungmin and Kalsi, Montek and Fox, Roy},
  journal={arXiv preprint arXiv:2403.12309},
  year={2024}
}

@article{lillicrap2015continuous,
  title={Continuous control with deep reinforcement learning},
  author={Lillicrap, Timothy P and Hunt, Jonathan J and Pritzel, Alexander and Heess, Nicolas and Erez, Tom and Tassa, Yuval and Silver, David and Wierstra, Daan},
  journal={arXiv preprint arXiv:1509.02971},
  year={2015}
}

@inproceedings{fujimoto2018addressing,
  title={Addressing function approximation error in actor-critic methods},
  author={Fujimoto, Scott and Hoof, Herke and Meger, David},
  booktitle={International conference on machine learning},
  pages={1587--1596},
  year={2018},
  organization={PMLR}
}

@inproceedings{haarnoja2018soft,
  title={Soft actor-critic: Off-policy maximum entropy deep reinforcement learning with a stochastic actor},
  author={Haarnoja, Tuomas and Zhou, Aurick and Abbeel, Pieter and Levine, Sergey},
  booktitle={International conference on machine learning},
  pages={1861--1870},
  year={2018},
  organization={Pmlr}
}

@article{bengio1994learning,
  title={Learning long-term dependencies with gradient descent is difficult},
  author={Bengio, Yoshua and Simard, Patrice and Frasconi, Paolo},
  journal={IEEE transactions on neural networks},
  volume={5},
  number={2},
  pages={157--166},
  year={1994},
  publisher={IEEE}
}

@article{wu2025directly,
  title={Directly forecasting belief for reinforcement learning with delays},
  author={Wu, Qingyuan and Wang, Yuhui and Zhan, Simon Sinong and Wang, Yixuan and Lin, Chung-Wei and Lv, Chen and Zhu, Qi and Schmidhuber, J{\"u}rgen and Huang, Chao},
  journal={arXiv preprint arXiv:2505.00546},
  year={2025}
}

@article{luo2024survey,
  title={A survey on model-based reinforcement learning},
  author={Luo, Fan-Ming and Xu, Tian and Lai, Hang and Chen, Xiong-Hui and Zhang, Weinan and Yu, Yang},
  journal={Science China Information Sciences},
  volume={67},
  number={2},
  pages={121101},
  year={2024},
  publisher={Springer}
}

@article{huang2022robust,
  title={Robust reinforcement learning as a stackelberg game via adaptively-regularized adversarial training},
  author={Huang, Peide and Xu, Mengdi and Fang, Fei and Zhao, Ding},
  journal={arXiv preprint arXiv:2202.09514},
  year={2022}
}

@article{miranda2023generalization,
  title={Generalization in deep reinforcement learning for robotic navigation by reward shaping},
  author={Miranda, Victor RF and Neto, Armando A and Freitas, Gustavo M and Mozelli, Leonardo A},
  journal={IEEE Transactions on Industrial Electronics},
  volume={71},
  number={6},
  pages={6013--6020},
  year={2023},
  publisher={IEEE}
}

@article{karamzade2025model,
  title={Model-Based Reinforcement Learning under Random Observation Delays},
  author={Karamzade, Armin and Kim, Kyungmin and Lanier, JB and Corsi, Davide and Fox, Roy},
  journal={arXiv preprint arXiv:2509.20869},
  year={2025}
}

@article{slaoui2019robust,
  title={Robust visual domain randomization for reinforcement learning},
  author={Slaoui, Reda Bahi and Clements, William R and Foerster, Jakob N and Toth, S{\'e}bastien},
  journal={arXiv preprint arXiv:1910.10537},
  year={2019}
}

@article{ajani2023evaluating,
  title={Evaluating domain randomization in deep reinforcement learning locomotion tasks},
  author={Ajani, Oladayo S and Hur, Sung-ho and Mallipeddi, Rammohan},
  journal={Mathematics},
  volume={11},
  number={23},
  pages={4744},
  year={2023},
  publisher={MDPI}
}

@inproceedings{zhao2021robust,
  title={Robust domain randomised reinforcement learning through peer-to-peer distillation},
  author={Zhao, Chenyang and Hospedales, Timothy},
  booktitle={Asian Conference on Machine Learning},
  pages={1237--1252},
  year={2021},
  organization={PMLR}
}

@article{morimoto2005robust,
  title={Robust reinforcement learning},
  author={Morimoto, Jun and Doya, Kenji},
  journal={Neural computation},
  volume={17},
  number={2},
  pages={335--359},
  year={2005},
  publisher={MIT Press}
}

@article{vinitsky2020robust,
  title={Robust reinforcement learning using adversarial populations},
  author={Vinitsky, Eugene and Du, Yuqing and Parvate, Kanaad and Jang, Kathy and Abbeel, Pieter and Bayen, Alexandre},
  journal={arXiv preprint arXiv:2008.01825},
  year={2020}
}

@article{pattanaik2017robust,
  title={Robust deep reinforcement learning with adversarial attacks},
  author={Pattanaik, Anay and Tang, Zhenyi and Liu, Shuijing and Bommannan, Gautham and Chowdhary, Girish},
  journal={arXiv preprint arXiv:1712.03632},
  year={2017}
}

@article{garcia2015comprehensive,
  title={A comprehensive survey on safe reinforcement learning},
  author={Garc{\i}a, Javier and Fern{\'a}ndez, Fernando},
  journal={Journal of Machine Learning Research},
  volume={16},
  number={1},
  pages={1437--1480},
  year={2015}
}

@inproceedings{achiam2017constrained,
  title={Constrained policy optimization},
  author={Achiam, Joshua and Held, David and Tamar, Aviv and Abbeel, Pieter},
  booktitle={International conference on machine learning},
  pages={22--31},
  year={2017},
  organization={PMLR}
}

@article{chow2018lyapunov,
  title={A lyapunov-based approach to safe reinforcement learning},
  author={Chow, Yinlam and Nachum, Ofir and Duenez-Guzman, Edgar and Ghavamzadeh, Mohammad},
  journal={Advances in neural information processing systems},
  volume={31},
  year={2018}
}

@article{dalal2018safe,
  title={Safe exploration in continuous action spaces},
  author={Dalal, Gal and Dvijotham, Krishnamurthy and Vecerik, Matej and Hester, Todd and Paduraru, Cosmin and Tassa, Yuval},
  journal={arXiv preprint arXiv:1801.08757},
  year={2018}
}

@inproceedings{pham2018optlayer,
  title={Optlayer-practical constrained optimization for deep reinforcement learning in the real world},
  author={Pham, Tu-Hoa and De Magistris, Giovanni and Tachibana, Ryuki},
  booktitle={IEEE International Conference on Robotics and Automation (ICRA)},
  pages={6236--6243},
  year={2018},
  organization={IEEE}
}

@inproceedings{ames2019control,
  title={Control barrier functions: Theory and applications},
  author={Ames, Aaron D and Coogan, Samuel and Egerstedt, Magnus and Notomista, Gennaro and Sreenath, Koushil and Tabuada, Paulo},
  booktitle={European control conference (ECC)},
  pages={3420--3431},
  year={2019},
  organization={Ieee}
}

@inproceedings{taylor2020learning,
  title={Learning for safety-critical control with control barrier functions},
  author={Taylor, Andrew and Singletary, Andrew and Yue, Yisong and Ames, Aaron},
  booktitle={Learning for dynamics and control},
  pages={708--717},
  year={2020},
  organization={PMLR}
}

@article{kushwaha2025survey,
  title={A Survey of Safe Reinforcement Learning and Constrained MDPs: A Technical Survey on Single-Agent and Multi-Agent Safety},
  author={Kushwaha, Ankita and Ravish, Kiran and Lamba, Preeti and Kumar, Pawan},
  journal={arXiv preprint arXiv:2505.17342},
  year={2025}
}

@article{neto2024robust,
  title={Robust longitudinal control for vehicular platoons using deep reinforcement learning},
  author={Neto, Armando Alves and Mozelli, Leonardo Amaral},
  journal={IEEE Transactions on Intelligent Transportation Systems},
  volume={25},
  number={10},
  pages={14401--14410},
  year={2024},
  publisher={IEEE}
}

@article{ray2019benchmarking,
  title={Benchmarking safe exploration in deep reinforcement learning},
  author={Ray, Alex and Achiam, Joshua and Amodei, Dario},
  journal={arXiv preprint arXiv:1910.01708},
  volume={7},
  number={1},
  pages={2},
  year={2019}
}

@article{towers2024gymnasium,
  title={Gymnasium: A standard interface for reinforcement learning environments},
  author={Towers, Mark and Kwiatkowski, Ariel and Terry, Jordan and Balis, John U and De Cola, Gianluca and Deleu, Tristan and Goul{\~a}o, Manuel and Kallinteris, Andreas and Krimmel, Markus and KG, Arjun and others},
  journal={arXiv preprint arXiv:2407.17032},
  year={2024}
}

@inproceedings{todorov2012mujoco,
  title={Mujoco: A physics engine for model-based control},
  author={Todorov, Emanuel and Erez, Tom and Tassa, Yuval},
  booktitle={IEEE/RSJ international conference on intelligent robots and systems},
  pages={5026--5033},
  year={2012},
  organization={IEEE}
}

@article{chen2024efficient,
  title={Efficient Reinforcement Learning With Impaired Observability: Learning to Act With Delayed and Missing State Observations},
  author={Chen, Minshuo and Meng, Jie and Bai, Yu and Ye, Yinyu and Poor, H Vincent and Wang, Mengdi},
  journal={IEEE Transactions on Information Theory},
  volume={70},
  number={10},
  pages={7251--7272},
  year={2024},
  publisher={IEEE}
}

@article{du2022random,
  title={Random-Delay-Corrected Deep Reinforcement Learning Framework for Real-World Online Closed-Loop Network Automation},
  author={Du, Keliang and Wang, Luhan and Liu, Yu and Niu, Haiwen and Huang, Shaoxin and Wen, Xiangming},
  journal={Applied Sciences},
  volume={12},
  number={23},
  pages={12297},
  year={2022},
  publisher={MDPI}
}

@article{janner2019trust,
  title={When to trust your model: Model-based policy optimization},
  author={Janner, Michael and Fu, Justin and Zhang, Marvin and Levine, Sergey},
  journal={Advances in neural information processing systems},
  volume={32},
  year={2019}
}

@article{hafner2019dream,
  title={Dream to control: Learning behaviors by latent imagination},
  author={Hafner, Danijar and Lillicrap, Timothy and Ba, Jimmy and Norouzi, Mohammad},
  journal={arXiv preprint arXiv:1912.01603},
  year={2019}
}

@article{moos2022robust,
  title={Robust reinforcement learning: A review of foundations and recent advances},
  author={Moos, Janosch and Hansel, Kay and Abdulsamad, Hany and Stark, Svenja and Clever, Debora and Peters, Jan},
  journal={Machine Learning and Knowledge Extraction},
  volume={4},
  number={1},
  pages={276--315},
  year={2022},
  publisher={MDPI}
}

@inproceedings{panaganti2022sample,
  title={Sample complexity of robust reinforcement learning with a generative model},
  author={Panaganti, Kishan and Kalathil, Dileep},
  booktitle={International Conference on Artificial Intelligence and Statistics},
  pages={9582--9602},
  year={2022},
  organization={PMLR}
}

@article{zhang2020robust,
  title={Robust multi-agent reinforcement learning with model uncertainty},
  author={Zhang, Kaiqing and Sun, Tao and Tao, Yunzhe and Genc, Sahika and Mallya, Sunil and Basar, Tamer},
  journal={Advances in neural information processing systems},
  volume={33},
  pages={10571--10583},
  year={2020}
}

@article{emam2022safe,
  title={Safe reinforcement learning using robust control barrier functions},
  author={Emam, Yousef and Notomista, Gennaro and Glotfelter, Paul and Kira, Zsolt and Egerstedt, Magnus},
  journal={IEEE Robotics and Automation Letters},
  year={2022},
  publisher={IEEE}
}

@article{zhang2021multi,
  title={Multi-agent reinforcement learning: A selective overview of theories and algorithms},
  author={Zhang, Kaiqing and Yang, Zhuoran and Ba{\c{s}}ar, Tamer},
  journal={Handbook of reinforcement learning and control},
  pages={321--384},
  year={2021},
  publisher={Springer}
}

@article{foerster2016learning,
  title={Learning to communicate with deep multi-agent reinforcement learning},
  author={Foerster, Jakob and Assael, Ioannis Alexandros and De Freitas, Nando and Whiteson, Shimon},
  journal={Advances in neural information processing systems},
  volume={29},
  year={2016}
}

@inproceedings{tobin2017domain,
  title={Domain randomization for transferring deep neural networks from simulation to the real world},
  author={Tobin, Josh and Fong, Rachel and Ray, Alex and Schneider, Jonas and Zaremba, Wojciech and Abbeel, Pieter},
  booktitle={IEEE/RSJ international conference on intelligent robots and systems (IROS)},
  pages={23--30},
  year={2017},
  organization={IEEE}
}

@inproceedings{chebotar2019closing,
  title={Closing the sim-to-real loop: Adapting simulation randomization with real world experience},
  author={Chebotar, Yevgen and Handa, Ankur and Makoviychuk, Viktor and Macklin, Miles and Issac, Jan and Ratliff, Nathan and Fox, Dieter},
  booktitle={International Conference on Robotics and Automation (ICRA)},
  pages={8973--8979},
  year={2019},
  organization={IEEE}
}

\end{document}